# A method for visuo-spatial classification of freehand shapes freely sketched

**Ney RENAU-FERRER, Céline REMI**
LAMIA, Université des Antilles et de la Guyane, Pointe à Pitre, Guadeloupe

***Abstract -*** *We present the principle and the main steps of a new method for the visuo-spatial analysis of geometrical sketches recorded online. Visuo-spatial analysis is a necessary step for multi-level analysis. Multi-level analysis simultaneously allows classification, comparison or clustering of the constituent parts of a pattern according to their visuo-spatial properties, their procedural strategies, their structural or temporal parameters, or any combination of two or more of those parameters. The first results provided by this method concern the comparison of sketches to some perfect patterns of simple geometrical figures and the measure of dissimilarity between real sketches. The mean rates of good decision higher than 95% obtained are promising in both cases.*

## 1 Introduction

When someone is freely drawing, an observer can simultaneously recognize the class of the final shape and evaluate its quality compared to a visual pattern. Thanks to a continuous observation, he can also capture each step which contributes to the building of the shape. This procedural analysis establishes the order according to which some predefined objects are drawn and the contribution of each recorded stroke (part of the sketch between a pen-up and a pen-down) to the emergence of each one of these objects. Furthermore, if he has appropriate models, the observer can assess the efficiency of the procedure in-progress. The four types of analysis described before are related to the visuo-spatial trace and the sketching activity. Someone can also analyse the handwriting movements produced and how the handwriting device is held. This fifth analysis is biomechanical. We name the joint realization of these five types of analysis during an act of producing a figure that multilevel analysis.

This paper is about a work that we have initiated on the problem of the simulation of this type of process. Our goal is the conception of tools to evaluate and assist the enhancement of graphomotor abilities and cognitive skills involved into efficient handwriting [1]. The paper is organised as follows.

The first section will specify why existing automated analysis methods do not satisfy our needs. Then, we will describe the common description that we have chosen to use for each process involved in a multilevel automated analysis of geometrical sketches. Among the five possible processes, we will focus on visuo-spatial classification. Our method will be exposed. Next, we will provide and discuss the results obtained on various types of sketches. Finally we will conclude our talk with some propositions to improve the performances of our system of visuo-spatial classification.

## 2 Previous Work

The pen-based interfaces coupled to softwares such as: EDT [2] or Anatrace [3], allow online acquisition of sketches. There are numerous fields for which methods of automated analysis of layouts were conceived. For example, we quote: assistance with the tracking or diagnostic of pathologies affecting graphomotricity or space perception [4], the discovery of strategies that children set in place to write [5]. The techniques of analysis at stake are generally conceived for a type or a specific set of figures and even for a particular writing task. So they tend to miss generics. Lastly, they do not allow at the very best two of the various levels of possible analysis. This is achieved with a strong propensity to give privilege to spatio-graphical analysis. Other works, with applications in fields such as electric engineering or training assistance to draw, generally deal with the analysis of handwritten layouts. Their aim is, for example, the computer-aided design of technical diagrams [6], [7], [8]. The final objective of these types of work is the visuo-spatial qualification of the treated forms. They are located for a number of them in interactive contexts of use. They are founded on the principle according to which: a feature/a temporal phase of the process of interaction gives place to the production of a particular object (electric component, left the face.). This destroys in fact the possibilities to analyze the layout activity itself.

So far as we know, there is no automatic technique of analysis simultaneously allowing the five aspects of a multi levels analysis mentioned above. The methods which are most similar to it, are intended in training educational uses such as those proposed in [9] or [10]. However, nature, the conditions, the order of production of the layouts and the mode of interaction they impose place them within the same a priori rigid framework as the others. This choice of their originators is justified by the fact that the type of script writer considered is a priori able to respect scriptural conventions to facilitate the automated analysis of its layouts. The effectiveness of the conceived system is to some extent based on the user keeping to the following principle: a produced feature = an object.

As far as we're concerned, the goal is to recognize the principle of construction of a figure implemented by a writer and compare it with principles of construction which are, for example, considered as the most efficient. Then, it is neither relevant to make the assumption that the writers are a priori able to conform to optimal or shared principles of construction, than they do. On the contrary, it is necessary to leave the principle that the writer is free to proceed throughout his production as he knows or can do it. As the production process progresses, the system of analysis has triple responsibility:

1. to perceive probable primitive entities within the traced features,

2. to extract useful information for each perceived primitive entity;

3. to refer each primitive entity perceived compared to the elementary geometrical objects awaited within the layout.

## 3 Adopted description

As far as we're concerned, the sketches are carried out with no procedural constraint of any kind. The elementary objects can thus be built in multiple strokes. Otherwise, the same stroke can contribute to the construction of several objects. The process of multilevel analysis needs a description halfway between rough description (made of sampled points) and the one based on strokes. Whatever field that we consider, the geometrical figures are generally visually perceptible as compositions of simple elements such as segments of right-hand side and arcs of circle. We thus chose to exploit these two types of primitives to manage the multilevel analysis of the layouts. We use the mixed technique of segmentation described in [11]. Unlike approaches mentioned in [5] and [12], this approach provides a probable parametric description while avoiding to take into account the a priori knowledge on nature of the elements composing the reproduced figures. Thus, this approach offers us considerable latitude in terms of diversity of the geometrical types of figures which can be treated.

During the segmentation, we label each sampled point as characteristic or not of the layout and pertaining to one of the identified primitives. In the same way, the primitives are labelled in comparison with the features which carry them.

There are various kinds of characteristic points. They are points of discontinuity of functions such as: curvature, speed, pressure, points of beginning and end of strokes, points of intersection of the strokes. The same sampled point can have several labels. According to the type of analysis considered, we'll only take into account the points that carry useful information for the analysis; they are then named: interest points.

## 4 Visuo-spatial classification

Visuo-spatial classification aims at establishing the degree of visual similarity of a layout acquired online with patterns. This has to be done, independently from their orientations and dimensions. The approach that we propose includes 3 steps. The first step consists in selecting interest points among the characteristic points. The second stage consists in computing a local descriptor of each one of these interest points. The last stage is classification itself, based on the estimation of the dissimilarity between a set of local descriptors describing a layout and a given pattern.

### 4.1 Selection of the interest points

For the visuo-spatial analysis, it is useless to code information in the neighbourhood of characteristic points spatially close to one another more than once [13]. However, if a writer passes by again his pen on a pre-existent feature (we call this phenomenon redrawing) this can generate more than one characteristic point for the same zone as illustrated on figure 1.

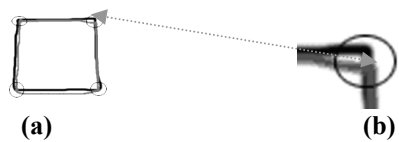

(a)                  (b)

**Figure 1 :: a sketch (a) and two characteristic points into a zone of redrawing (b)**

It's also possible to find close characteristic points belonging to primitives which contribute to form a pattern of known geometrical object. We call them Structural Neighbours. When we detect characteristic points that are grouped in a small zone where the writer went back over a drawing, or structural neighbours, we take into account the last point that was recorded.

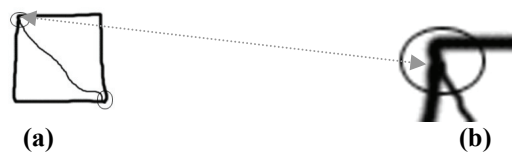

(a)                  (b)

**Figure 2 : a sketch (a) and an example of structural neighbours (b)**

## 4.2 Local descriptors

A local descriptor is defined by the spatial distribution of the points of the layout in the neighbourhood of each interest point. As for the neighbourhood, we chose to consider the circular area around the interest point. This circular area will be divided into 16 portions of 22.5° each in order to deduce a histogram with fixed step from it. The points of the layout taken into account are those which are contained in the area of the descriptor and which belong to a primitive starting with, ending in, or passing by a characteristic point located in the area of the descriptor. We standardize the number of the points contained in each portion of a descriptor by taking into account both the length of the smallest implied primitive and speed-writing variability. It is extremely important to pay the greatest attention to the initial adjustment of the origin for the descriptor. The histograms can be disturbed because of the "overflow" of the points describing a primitive in close portions. To limit the impact of these border effects, first, we determine a principal axis of inertia to position the descriptor. Then, as soon as each portion of the descriptor makes 22.5°, we carry out a series of rotations of 1° going from -11° to +11° and preserve the positioning which gives rise to the histogram presenting the greatest standard deviation. This procedure provides histograms clearer as one can note on figure 3c.

$$D(U,V) = \sum_{i=1}^{16} K * \frac{(u_i - v_i)^2}{u_i + v_i} \quad (1)$$

Where : U and V are 2 descriptors.

$u_I$ and $v_I$ are the number of points for U and V in $I^{ème}$ portion of the descriptor.

K is worth 1 plus the number of portions between $I^{ème}$ peak of U and $I^{ème}$ peak of V

If a portion of layout is symmetrical to another by symmetry of axis X or the origin of a reference mark, one cannot match the other by a rotation. To mitigate this problem we number the portions of the descriptor in the positive OR negative direction according to the sign of the slope coefficient of the axis of inertia. Lastly, when comparing the two descriptors we make the first one make 16 rotations of 22.5° and compute a distance for each position. We finally preserve the smallest distance. A portion of the descriptor is regarded as a peak if its number of points is higher than the average density of the nonempty peaks.

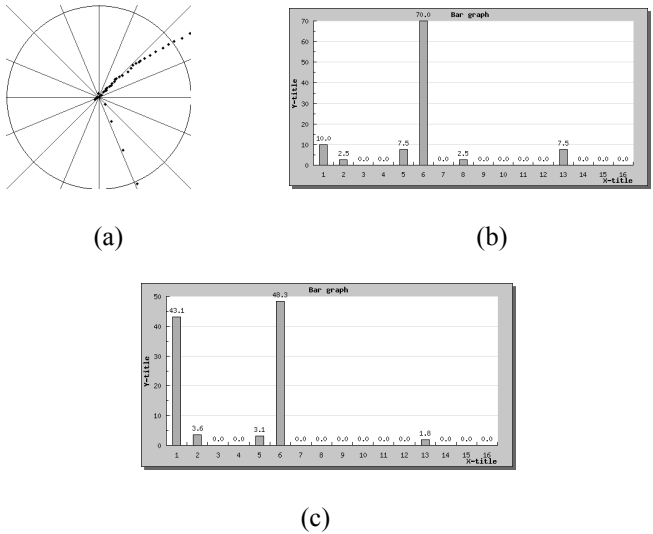

**Figure 3: Histograms without (b) and with (c) weighting for a interest point selected on a layout (a)**

## 4.3 Dissimilarity measure

We want to take simultaneously into account the difference between densities of points as well as the angular divergence between the portions of layouts observed. With this aim , we use the measure suggested in [14] to which we integrate a parameter K.

## 5 Experimental results

### 5.1 Test on synthetic sketches

In order to judge the relevance of our descriptor and that of our dissimilarity measure, we choose to test the method on a set of synthetic geometrical sketches. Synthetic sketches are interesting because they free the method from inherent difficulties of the real conditions of freehand sketch like: local distortions of strokes, numerous redrawingr and structural neighbours. Our set contains 18 synthetic sketches in 5 classes illustrated on the figure 4.

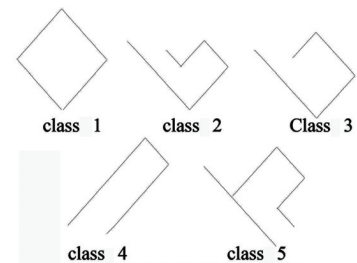

**Figure 4 : synthetic patterns for 5 classes used to test the local descriptor and the dissimilarity measure**

These figures were generated by respecting some constraints to facilitate any other treatment than the generation of the descriptor and the estimation of the dissimilarity.

The classification accuracy rate is 100% for the synthetic sketches.

This result validates the principle of the descriptor and the distance used for the estimation of the visuo-spatial dissimilarity.

## 5.2 Results on real sketches

We have to study the behaviour of our method on real freehand sketches. For this purpose, we chose a subset of the HHreco dataset [15]. This subset contains the sketches produced by 19 different users. Each user had drawn at least 30 examples for each of the 5 classes which are presented on figure 5.

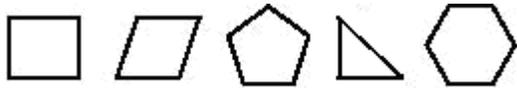

**Figure 5: 5 classes of the HHreco dataset used to test the method on real sketches**

As shown on figure 6, the users have reproduced the patterns with one or more strokes. Sometimes, their sketches present numerous redrawings, points of intersection and unfinished forms (i.e. : open).

3023 sketches were classified according to their similarity with "perfect" patterns representing each class.

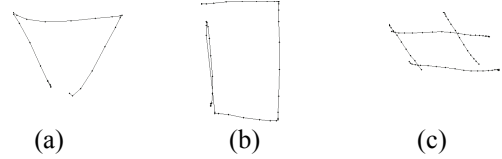

(a)  (b)  (c)

**Figure 6: open triangles (a), square with a redrawing zone (b) & parallelogram with secant strokes (c)**

By "perfect" we mean that the pattern doesn't present redrawing or structural neighbours. Each one of the "perfect" patterns chosen meets one of the following predefined grapho-perceptive requirements: a perfect square has only 4 angles of 90°, a perfect triangle is an equilateral right-angled triangle, a perfect parallelogram is a parallelogram with 2 angles of 45° and 2 angles of 135°, a perfect pentagon is composed of two right angles, two angles of 112,5° and an angle of 135°, the hexagon is similar to the pentagon but with an additional angle of 135°.

| Recognition rates for : | Triangle | Square | Parallelogram | Pentagon | Hexagon | Global rate |
|---|---|---|---|---|---|---|
| user 01 | 93,33% | 93% | 93,33% | 86,66% | 100% | 93,24% |
| user 02 | 83,33% | 70,00% | 40,00% | 90,32 | 100% | 76,82% |
| user 03 | 96,66% | 93,30% | 96,66% | 80% | 100% | 93,33% |
| user 04 | 96,66% | 76,66% | 93,33% | 90% | 100% | 91,33% |
| user 05 | 96,77% | 88,87% | 60% | 93,54% | 100% | 86,92% |
| user 06 | 100% | 100% | 100% | 90% | 100% | 98% |
| user 07 | 100% | 96,66% | 86,66% | 96,66% | 100% | 96,10% |
| user 08 | 83,33% | 96,66% | 86,66% | 100% | 100% | 93,33% |
| user 09 | 100% | 100% | 80% | 90% | 100% | 94% |
| user 10 | 100% | 100% | 66,66% | 96,66% | 100% | 92,66% |
| user11 | 96,66% | 100% | 96,66% | 87% | 96,77% | 95,39% |
| user12 | 100% | 96,66% | 100% | 90% | 100% | 97,33% |
| user13 | 96,66% | 96,66% | 86,66% | 90% | 100% | 94,07% |
| user14 | 100% | 96,66% | 93,33% | 90% | 100% | 96% |
| user15 | 81,63% | 94% | 96% | 90% | 100% | 91,96% |
| user16 | 100% | 96,96% | 93,94% | 88,24% | 100% | 95,70% |
| user17 | 87,09% | 93,33% | 73,33% | 96,66% | 100% | 90% |
| user18 | 96,66% | 96,66% | 93,33% | 86,66% | 96,66% | 94% |
| user19 | 100% | 96,66% | 70,00% | 93,33% | 100% | 92% |

**Table 1: recognition rates by user with once perfect pattern for each class**

The rates of correct classification of sketches by user are shown in Table 1. The rate of good recognition is the lowest for the parallelograms whereas that of the squares is the 2nd higher rank. It is simply explained by the fact that when drawing a square, there's little variability since there is only one possible ideal square (4 right angles) whereas, there is a large variety of patterns when drawing a parallelogram. Moreover, the hexagon shows the best results whereas the pentagon is next to last. This can be explained by the fact that the pentagon is a class which one could describe as median. So, the system can sometimes hesitate between pentagon and hexagon or between pentagon and square whereas the hexagon can only be mistaken for the pentagon. Although, the rate of global recognition is considerable, it remains improvable by increasing the capacity of good recognition for classes like pentagon and parallelogram.

Thanks to our method of comparison to standard patterns, we intended to make a layout cluster around the most similar single model for each class.

But, there are an infinite number of triangles, pentagons, hexagons, parallelograms. This infinity can be brought back to a finished number owing to the fact that our descriptor does not see the variations which are lower than 22,5°. The flexibility of the method allows, for a given class, the comparison to several patterns instead of one. We tested this principle on each group of layouts presenting a percentage of recognition lower than 90%. This contributed significantly to improve the rates of good recognition for these classes as shown on the second row of table 2.

Let us note that the application of this method only on the classes which had obtained less than 90% increased the global rate of recognition to 95%.

The last line of the table 2 establishes that in the case of comparisons with freehand sketches taken as references, the method of classification also provides high rates of good decisions.

|  | Triangle | Square | Parallelogram | Pentagon | Hexagon | Global rate |
|---|---|---|---|---|---|---|
| **Global rate with once perfect pattern used as a single model for each class** | 94,68% | 93,69% | 84,90% | 90,77% | 99,67% | 92,75% |
| **Global progression observed for the users whose rates of recognition are the smaller ones (grey) when several patterns are used as model for each class** | +12,90% | +14,30% | +15,26% | +9,77% | - | +13,24% |
| **Global rate with freehand sketches as reference for each class** | 98,50% | 94,85% | 90,38% | 97% | 99,83% | 95,96% |

**Table 2: global rates of good recognition for each class according to different modalities.**

# 6  Conclusion

We have started a work about the simulation of human multi-levels analysis of sketching.

This paper focused on the method that we proposed for the visuo-spatial aspect of an automatic multi-levels analysis. Its purpose is to provide a relevant way to obtain the recognition and the qualitative evaluation of freely drawn sketches. First, a segmentation step provides the set of characteristic points. Secondly, interest points for the visuo-spatial discrimination of shapes are selected. Next, local descriptor are computed for each of these interest points which are not labelled as structural neighbour or redrawn. Finally, the KX distance is estimated between pairs of local descriptors and added to estimate the dissimilarity between shapes.

Three different tests were performed. The one performed on synthetic sketches gave 100% of good classification of shapes. It validates the principle of the local descriptor and the KX distance that we proposed. Another test performed with real sketches showed good behaviours in case of evaluation of visuo-spatial similarity between freehand shapes. The third type of test showed good classification rates in case of comparison between "perfect" patterns of shapes and sketches. However, due to the great variability between freehand figures into a class and the multiplicity of potential "perfect" patterns, results observed on certain classes have to be enhanced. Preliminary tests showed that the fact of considering several patterns for one class improved the rates of good decision for median classes. According to the field of application of the method, those tracks should be examined in more details.

Near this potential source of enhancement of our method, we will explore other tracks. For example, a more reliable phase of detection of the characteristic points and more particularly of the points of intersection could make it possible to improve the rates of recognition observed on the layouts which include secant strokes. Moreover, we intend to use the results of the phase of procedural analysis, to proceed to a feedback on the visuo-spatial analysis. The purpose of this feedback would be the correction of the results for the numerous cases where the system does not manage to choose the true class but, knowing that there's a tiny difference between the distance at the true class and the one to the selected class.